\documentclass[11pt,a4paper]{article}
\usepackage{times,latexsym}
\usepackage{rotating}
\usepackage{float}
\usepackage{url}
\usepackage[T1]{fontenc}
\usepackage[table,xcdraw]{xcolor}
\usepackage{tabularx}
\usepackage{makecell}
\usepackage{lscape}
\usepackage{longtable}
\usepackage{caption}
\usepackage{graphicx}
\usepackage{lscape}
\usepackage{multirow}
\usepackage{array}
\usepackage{times}
\usepackage{latexsym}
\usepackage[export]{adjustbox}

\usepackage[T1]{fontenc}

\usepackage[utf8]{inputenc}

\usepackage{microtype}

\usepackage{inconsolata}

\usepackage{graphicx}

\usepackage{array}
\usepackage{longtable}
\usepackage{geometry}
\usepackage{caption}
\usepackage{enumitem}
\usepackage{url}
\usepackage{hyperref} 

\usepackage[acceptedWithA]{tacl2021v1}
\usepackage{xspace,mfirstuc,tabulary}

\newif\iftaclinstructions
\taclinstructionsfalse 
\iftaclinstructions
\renewcommand{\confidential}{}
\renewcommand{\anonsubtext}{(No author info supplied here, for consistency with
TACL-submission anonymization requirements)}
\newcommand{\instr}
\fi

\iftaclpubformat 

\else

\fi

\title{A Systematic Review of NLP for Dementia- \\Tasks, Datasets and Opportunities}

\author{
    Lotem Peled-Cohen \and Roi Reichart \\
    Technion - Israel Institute of Technology \\
    \texttt{lotemi.peled@gmail.com} , \texttt{roireichart@gmail.com}
}

\date{}

\begin{document}
\maketitle
\begin{abstract}
The close link between cognitive decline and language has fostered long-standing collaboration between the NLP and medical communities in dementia research. To examine this, we reviewed over 240 papers applying NLP to dementia-related efforts, drawing from medical, technological, and NLP-focused literature. We identify key research areas, including dementia detection, linguistic biomarker extraction, caregiver support, and patient assistance, showing that half of all papers focus solely on dementia detection using clinical data. Yet, many directions remain unexplored- artificially degraded language models, synthetic data, digital twins, and more. We highlight gaps and opportunities around trust, scientific rigor, applicability and cross-community collaboration. We raise ethical dilemmas in the field, and highlight the diverse datasets encountered throughout our review- recorded, written, structured, spontaneous, synthetic, clinical, social media-based, and more. This review aims to inspire more creative, impactful, and rigorous research on NLP for dementia. 
\end{abstract}

\section{Introduction}
\label{sec:intro}


Dementia is a broad term for a decline in cognitive function caused by various underlying pathologies. It is a progressive, irreversible condition that worsens over time, with no known treatment or cure. The global impact of the disease is staggering. According to the World Health Organization (WHO),\textsuperscript{1} dementia is currently \textit{the seventh leading cause of death globally}. As of 2023, approximately 55 million people worldwide are living with dementia, and this number is expected to nearly double every 20 years \cite{prince2015world}. Global dementia costs are estimated in the trillions of US dollars, with approximately half of these costs attributed to care provided by informal carers (e.g., family members and close friends), as hospitals and care facilities are overcrowded, and healthcare professionals require specialized training to diagnose patients and address their complex needs.

There are multiple types of dementia, including Alzheimer’s (accounting for 60\%-70\% of cases\footnote{https://www.who.int/news-room/fact-sheets/detail/dementia}), vascular dementia, Lewy-body dementia, and others. Although these differ in pathology - some manifesting as abnormal protein deposits and others as reduced blood flow to the brain - they often share common symptoms such as memory loss, confusion, behavioral changes, and language deterioration \cite{mckhann2011diagnosis}. In this review, we use the general term `dementia' to encompass all forms and levels of severity of the illness.

Dementia is currently diagnosed through pathological tests (e.g., brain imaging, blood tests) and cognitive assessments, such as \textit{Naming and Verbal Fluency tests}, which assess the ability to identify objects or generate words in specific categories \cite{kaplan2001boston}, and \textit{Picture Description tests}, evaluating language and narrative skills through descriptions of complex images \cite{mueller2018connected}. In these face-to-face interviews, clinicians seek linguistic markers of cognitive decline, such as repetitive language, empty or disorganized speech, word-retrieval difficulties, and excessive descriptions ("the thing you write with" instead of "pen").

Given the prominent role of language in our understanding and diagnosis of the disease, it is no surprise NLP methodologies are widely used in both the NLP and medical communities. Cognitive assessments and similar diagnostics often result in recorded and transcribed data, offering a wealth of textual content that is well-suited for NLP analysis. This, combined with relatively new sources of data (such as social media posts or conversations with LLMs), enhance the potential of NLP to advance dementia research in various directions.

Previous literature reviews on NLP for dementia focus on detection methods \citep{clarke2020things, petti2020systematic, saleem2022deep, parsapoor2023ai, qi2023noninvasive, hiremath2023early, vrindha2023review}, with some specifically covering deep learning approaches to detection \citep{shi2023speech, javeed2023machine}. Other reviews extend beyond detection, addressing tasks like linguistic biomarker extraction \citep{gagliardi2024natural} or exploring the role of technology in patients' life \citep{d2017information, shahapure2022nlp, 
saragih2023effects, peres2024systematic}, though these are not fully focused on NLP methods. A few reviews are even more specific, covering dementia-related data \citep{mueller2018connected, de2020artificial, yang2022deep}. 

Our review stands out by covering the full spectrum of dementia-related NLP efforts, rather than focusing on specific aspects like tasks (e.g., detection), technologies (e.g., deep learning), data types (e.g., clinical), or communities (NLP vs.\ medical). Moreover, we created our review with NLP readers in mind, unlike prior reviews, which were primarily published in medical literature and aimed at clinicians (Figure~\ref{fig:dist_task_venue}). Additionally, our work is distinguished by its scope and coverage- we review \textit{242 papers} from four distinct scientific communities, categorized by publication venue: the medical community (e.g., Nature Medicine); the NLP community (e.g., ACL); the Speech community (e.g., Interspeech); and a broader technological community that is not necessarily NLP- or Speech-specific (e.g., Frontiers in Computer Science). In Section~\ref{sec:methodology}, we detail the construction of our cohort.

In Section~\ref{sec:task_families}, we describe the main task families identified in our reviewed papers: \textit{linguistic biomarker extraction, caregiver support, patient assistance}, and, most prominently, \textit{dementia detection}, which accounts for over 56\% of the cohort's focus. For each task family, we review motivations, current approaches, and potential future directions.

Section~\ref{sec:future} examines gaps and opportunities. We highlight, for instance, that the vast majority of studies - across all four task families - rely on a handful of well-known datasets, despite the existence of many unique datasets varying in size, type, and purpose (a summary of 17 dementia-specific datasets can be found in Table~\ref{tab:my-table}). We then dive into the field's scientific rigor and explores NLP's potential to influence medical research- provided trust between the two 
communities is established. 

To provide readers with a concrete takeaway on future research directions, Section 5 outlines open challenges such as personalized LLMs for patients and caregivers, artificially degraded language models and many more. We conclude with the unique ethical considerations of the field (Section 6) and the limitations of our review (Section 7).

Our work aims to inspire researchers from various fields to tap into the vast potential of NLP in dementia research. We hope to provide a fresh perspective on the domain, emphasizing that the opportunities extend far beyond detection. Whether developing clinical applications, analyzing disease progression, or alleviating caregiver burden, this review serves as a valuable resource for those looking to contribute to the fight against dementia.

\section{Methodology}
\label{sec:methodology}

This review was conducted according to the PRISMA guidelines \cite{moher2009preferred}. We searched for NLP- and dementia-related papers in titles, abstracts, and keywords across ACL Anthology, PubMed, DBLP, IEEE Xplore, Springer, and Wiley (full query details in Appendix~\ref{cohort}). An automatic screening ensured papers were full academic studies (rather than posters or theses), peer-reviewed, and in English. We then manually screened for eligibility, selecting studies that (1) focus on text as a primary modality, (2) use datasets that at least partially in English (to narrow our scope while still addressing multilingual studies), and (3) were relevant to this review. For instance, while \citet{botros2020simple} mentions dementia and language models in its abstract and keywords, it focuses on smart home sensors, making it ineligible. Appendix~\ref{cohort} provides further examples of papers that do not meet our relevance criteria, alongside a fully detailed overview of our query, screening process, inclusion criteria, and a PRISMA flowchart.

Our screening resulted in \textbf{242 relevant papers}, which we then manually annotated with their main contributions, datasets used, algorithmic methods and whether statistical significance was reported (where applicable). Among this cohort, we identified four distinct task families, described below.

\section{Task Families}
\label{sec:task_families}

\begin{figure}
    \centering
\includegraphics[width=0.8\linewidth]{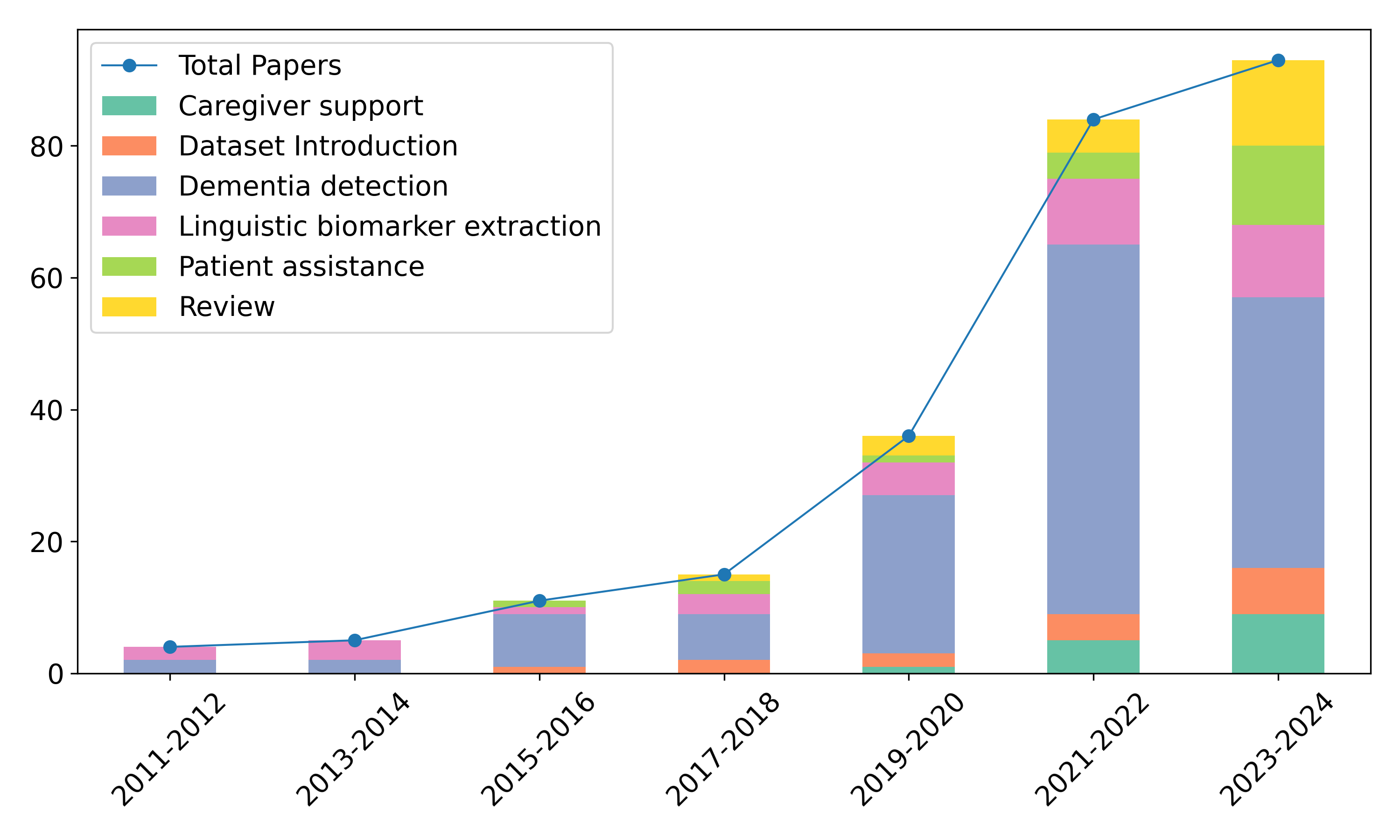}
    \caption{Number of papers per period by task family.}
    \label{fig:yearly}
    \includegraphics[width=0.8\linewidth]{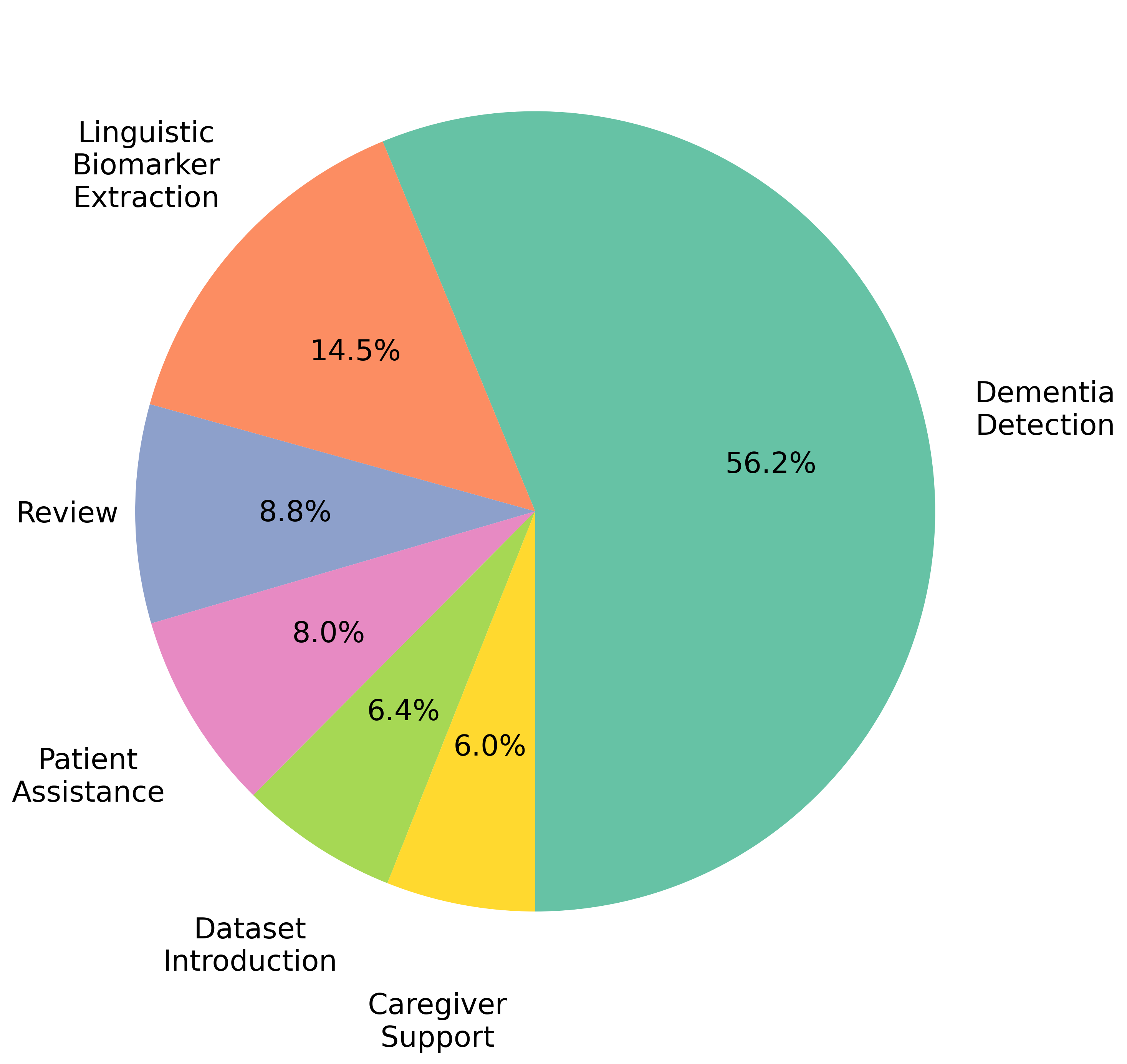}
    \caption{Distribution of task families across papers.}
    \label{fig:dist_sections}
    \vspace{-1em}
\end{figure}

We identified four distinct research verticals differing by tasks and motivations- \textbf{dementia detection, linguistic biomarker extraction, caregiver support} and \textbf{patient assistance} (see Section~\ref{sec:task_families}). Two additional, more general categories are \textbf{literature reviews} and \textbf{dataset introduction} papers, which we describe throughout our work. The number of studies on NLP for dementia has been steadily growing  (Figure~\ref{fig:yearly}), reflecting a significant interest in the field. While a rise in studies on patient assistance, caregiver support, and literature reviews, there is still an overwhelming focus on dementia detection (Figure~\ref{fig:dist_sections}). Naturally, this imbalance between the four tasks is reflected in our cohort - over 130 dementia detection papers versus 15-60 papers for each other task family - affecting the depth of our analysis of specific dementia detection studies. Therefore, Appendix~\ref{extended_citations} offers further details on the dementia detection papers we reviewed.

\paragraph{Dementia Detection}\label{detection}

Can an algorithm accurately predict dementia from a provided text? Remarkably, 56\% of the papers we reviewed address this exact question. This focus stems from two key factors. First, a \textit{tangible impact}: NLP can improve the diagnostic process by making it faster, less invasive, and more affordable. Pathological changes can begin 15-20 years before cognitive symptoms are noticeable to others \cite{jack2010hypothetical}, and if NLP algorithms can detect early stages of dementia - e.g., mild cognitive impairment (MCI) - they may enable earlier intervention, potentially slowing disease progression and delaying costly care.

The second motivation to pursue dementia detection is a \textit{data-driven} one. NLP research is inherently empirical, and in this domain, data is often structured - spanning medical, audio, or textual modalities - annotated with categorical labels such as `healthy' or `dementia.' This structure perfectly suits classification algorithms, making dementia detection a well-defined challenge for researchers.

NLP-based dementia detection has been an active research domain for over a decade, primarily leveraging well-known datasets such as the Pitt corpus \cite{becker1994natural}, part of the DementiaBank cohort \cite{lanzi2023dementiabank}. The Pitt corpus provides transcribed recordings of cognitive assessments, such as picture descriptions. These transcribed text excerpts (e.g., `there's a young boy, uh, going in a cookie jar, and there's a lit... a girl...') are annotated with the speaker's cognitive state (e.g., Healthy, Mild Cognitive Impairment, Alzheimer's) allowing for text-based classification. The transcriptions are also annotated with demographic information and, for some participants, other cognitive assessment scores. Two other widely used data sources, the ADReSS and ADReSSo challenge datasets \citep[respectively]{luz2021detecting, luz2021alzheimer}, are derivatives of the Pitt corpus, offering refined transcripts and a more demographically balanced sample. 

Unlike these datasets, which build on structured conversations, the widely used CCC dataset \cite{pope2011finding} contains transcriptions of spontaneous conversations, about memories, health, and daily life (e.g., `hmm. This is my problem (...) maybe you could help me find it'). These four popular datasets (Pitt, ADReSS, ADReSSo and CCC) exemplify the most common data acquisition approach in dementia detection: transcribed conversations (whether from well-known datasets or self-collected clinical data) annotated with Dementia or Healthy Control labels.\footnote{Some datasets offer more granular annotations, such as Dementia, Mild Cognitive Impairment, Alzheimer's, etc.}

Across the 136 detection papers we reviewed, several algorithmic approaches emerged. Please note that we present representative citations here; for the complete list of references, see Appendix~\ref{extended_citations}. Before 2017, many studies relied on classic machine learning methods like SVM and Random Forests using straightforward features like N-grams \citep{jarrold2014aided, fraser2016linguistic, zhou2016speech, santos2017enriching}, achieving accuracy rates around 85\%. From 2018 onwards, there was a surge in transformer-based classifiers, particularly applied to the ADReSS and ADReSSo challenges published around that time \citep{pappagari2020using, edwards2020multiscale, haulcy2021classifying, balagopalan2021comparing}. These methods pushed accuracy over 90\%, which are currently state-of-the-art results on the Pitt corpus and its subsets. Recently, large language models (LLMs) have been utilized for detection via prompt design, embedding extraction, and feature design \citep{agbavor2022predicting, wang2023text, runde2024optimization, bt2024performance, bang2024alzheimer}. 

Given the straightforward nature of the dementia detection (a text classification task), extensive body of research, and the impressive classification results, one might assume that dementia detection is a solved problem. From a modeling perspective, it has indeed evolved like other text classification tasks - from classic methods to word embeddings, neural networks, and LLMs. However, it remains fundamentally different, requiring tailored preprocessing, posing challenges in data collection from older adults, and demanding specialized linguistic metrics for evaluation - complexities we explore further in our review. As for the truly impressive results, they are (a) largely based on the Pitt corpus or its subsets - small, homogeneous datasets with specific characteristics (structured conversation, English only, etc.); (b) only 29\% of studies report statistical significance or assess the robustness of their findings; and (c) despite strong performance, no NLP-based classification tool, to our knowledge, has been deployed in real-world applications. We explore these gaps further in Section~\ref{sec:future}.

\vspace{-0.2em}
\paragraph{Linguistic Bio-Markers Extraction}

Some studies leverage NLP for linguistic exploration, rather than straightforward detection. They often wish to confirm, on a large scale, the existing knowledge we have about the language of cognitively impaired individuals. For instance, studies such as \citet{orimaye2014learning}, \citet{rosas2019search} and \citet{ilias2022explainable} used NLP methods to verify that linguistic cues associated with cognitive decline - such as repetitions (e.g., `the... the...'), revisions (e.g., `the woman, uh, mother'), and overuse of pronouns (e.g. `she' instead of `mother') - are indeed behaviors significantly more common in texts produced by the cognitively impaired \citep{clark2014lexical, fraser2016linguistic, voleti2019review}.

Other studies introduce linguistic markers and metrics to assess a speaker's cognitive state. For example, \citet{roark2007syntactic} extracted \textit{complexity scores} from parse trees, quantifying the extent to which sentences produced by cognitively impaired individuals are structurally and grammatically less complex. \citet{sirts2017idea} calculated idea density to measure how efficiently dementia patients convey ideas, while \citet{pompili2020pragmatic} quantified the number and order of topics mentioned by speakers in cognitive interviews. In other studies, such as \citet{choi2019meta}, the use of \textit{meta-semantic terms} - words implying emotion, emphasis, or opinion - was analyzed in picture descriptions as an indicator of cognitive performance. The \textit{frequency of disfluencies}, including silent pauses, reformulations, and context switches \citep{adhikari2021comparative, farzana2022you, williamsinhibitory}, has also been identified as a strong linguistic marker, even enabling the longitudinal tracking of disease progression \citep{martinc2021temporal, robin2023automated}. 

Findings from this task family impact both the fields of NLP and medicine. From an NLP perspective, they underscore that the language of cognitively impaired individuals differs drastically from typical language, presenting unique challenges to standard NLP practices. For example, these studies show that certain preprocessing steps traditionally used in NLP, such as removing repetitions and stop words, could inadvertently erase dementia-related linguistic signals that are critical for tasks like dementia detection. From a medical perspective, these studies have the potential to challenge conventional scoring practices in cognitive assessments. One example is \citet{prud2011using}, who explored narrative recall tasks - where participants are asked to retell a story after hearing it. The researchers found that scoring this cognitive assessment based on the presence of \textit{specific story elements} may be more effective for detecting MCI than scoring based on \textit{the overall summary}, which is the traditional approach, thus challenging the current perspective on this task. 

Notably, 75\% of papers in this task family report statistical significance- the highest ratio among all task families. This likely stems from the motivation to ensure that existing knowledge, or any new linguistic insights, are robust.

\vspace{-0.2em}
\paragraph{Caregiver Support} 

With the growing number of dementia patients, the demand for caregivers rises. In the US alone, an estimated 11 million Americans serve as caregivers for those with dementia, contributing billions of hours of care each year.\footnote{https://www.alzint.org/about/dementia-facts-figures/} Research indicates that nearly half of these caregivers experience depression, and are at a higher risk of chronic health conditions \cite{huang2022depression}. NLP methods can support these devoted caregivers by detecting emotional distress, providing answers to their concerns and offering companionship.

Around 6\% of our cohort consists of studies on NLP for caregiver support, a field that emerged post-COVID-19. Early research focused on the emotional well-being of personal caregivers - family and friends - mainly through social media posts \citep{monfared2021understanding, azizi2024identifying}. For example, \citet{sunmoo2022analyzing, sunmoo2023comparing} used graph-based topic modeling and sentiment analysis to show that tweets posted throughout the pandemic shifted from practical care to emotional distress (e.g., depression, helplessness, elder abuse) and coping strategies (e.g., therapeutic reading). Posts beyond dementia-specific communities \citep{ni2022rough, lal2023hybrid} and across social media profiles \citep{klein2022automatically} reveal that caregivers express more than emotional distress, asking also for financial aid and legal advice.

Other studies focus on \textit{professional caregivers}. One example is \citet{zhu2022agitation}, who analyzed clinical notes by nurses in aged care facilities. Disturbingly, their analysis shows dozens of distinct aggressive behaviors targeted towards the nursing team - including pushing, shouting, and using profane language - involving over 50\% of 
dementia patients studied. This line of research highlights the potential of NLP in detecting when to assist caregivers, whether through advice, emotional support, or enhanced workforce training.

Another growing body of research is exploring whether LLMs could provide caregivers with expert answers to dementia-related inquiries. Current studies show that models like ChatGPT offer relevant and factually correct advice and recommendations \citep{hristidis2023chatgpt, dosso2024does}, but often lack the depth of information available through Google Search or expert sources. LLMs also fall short in offering the emotional support needed when a caregiver seeks to manage a loved one's memory loss, confusion, or aggression  \citep{aguirre2024assessing}. To bridge these gaps, \citet{zaman2023empowering} and \citet{parmanto2024reliable} explore the novel concept of fine-tuning LLMs to specifically address caregivers' delicate emotional needs. These studies pave the way for AI-based caregiver companions, a field likely to gain traction with the growing popularity of LLMs. While groundbreaking, these agent-based solutions also present critical ethical dilemmas, explored in Section 6.

\vspace{-0.3em}
\paragraph{Patient Assistance} 

Individuals with dementia can live with the disease for decades,\footnote{https://www.alzheimers.org.uk/about-dementia/symptoms-and-diagnosis/how-dementia-progresses/later-stages-dementia} and NLP solutions can significantly improve their daily lives. NLP can also help researchers decode the mental landscapes of patients, where agitation, depression, and cognitive decline intersect. Research indicates that depression affects 30-50\% of dementia patients \cite{lyketsos2003diagnosis}, with 15\% experiencing suicidal thoughts \cite{naismith2022suicidal}. This highlights the importance of studies such as \citet{fraser2016detecting} and \citet{ehghaghi2022data}, who aim to detect depression among dementia patients. These studies demonstrate that it is extremely challenging to detect depression from the transcribed cognitive assessments alone, and that adding multi-modal data, particularly audio, drastically helps.

\begin{figure*}[h]
    \centering
    \includegraphics[width=0.8\linewidth]{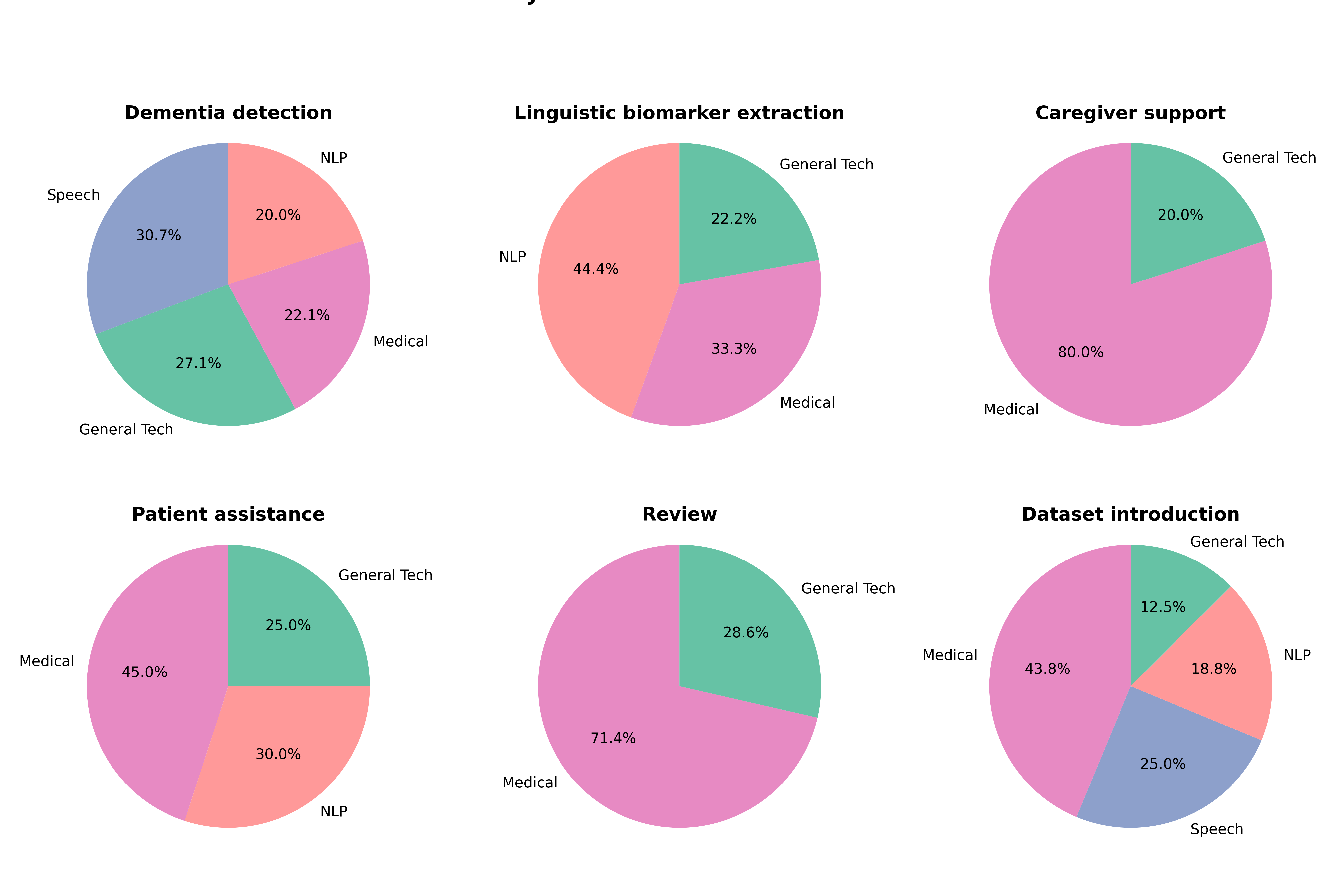}
    \caption{Distribution of venues across each of the task families.}
    \label{fig:dist_task_venue}
    \vspace{-1em}
\end{figure*}

Apart from assessing the mental well-being of patients, NLP can also aid in text simplification, making online information more accessible to those who often find it unreadable \cite{espinosa2023review}. \citet{engineer2023investigating} analyzed dementia-related texts online, finding an average readability level equivalent to 12 years of education - potentially inaccessible to many at-risk communities, given the link between lower education and dementia susceptibility.\footnote{https://www.ncbi.nlm.nih.gov/pmc/articles/PMC6937498/} The study also reveals a consistently negative tone in the texts (e.g., emphasizing the grim nature of the disease), potentially affecting the mental state of patients who are already prone to depression. NLP tools are natural candidates to help simplify texts, enhance readability, and rephrase to a more supportive tone. They can also assist writers and content creators ensure their work does not promote or reinforce stigma against patients \cite{pilozzi2020overcoming}.

Researchers are also trying to help patients sustain meaningful conversations and prevent communication breakdowns, shown to occur in over a third of interactions involving individuals with dementia.  Studies such as \citet{green2012assisting} and \citet{chinaei2017identifying} have developed systems to detect confusion and disfluent speech patterns within transcribed interviews, suggesting real-time repair strategies to minimize misunderstandings. Such solutions may enhance conversations for dementia patients while also easing the burden on caregivers.

A growing body of research explores LLM-powered chatbots to enhance social engagement, provide cognitive stimulation, and reduce loneliness \citep{kostis2022ai, xygkou2024mindtalker, qi2023chatgpt, gholizadeh2023conversational}. Studies show that participants appreciate the novelty and cognitive stimulation, reporting reduced loneliness and increased social support. However, they note some limitations such as handling emotionally sensitive conversations. Encouragingly, studies like \citet{addlesee2024you} focus on improving chatbot patience and empathy, advancing the development of more suitable LLMs for patients. 

Additionally, as with all AI tools, building trust with patients and caregivers is essential. \citet{gilman2024training} proposes specialized GPT training for dementia patients as a way to help establish this trust. Studies such as \citet{pacheco2024will} and \citet{treder2024introduction} interviewed patients, healthy individuals, and caregivers about their views on such technologies, revealing concerns about bias, data privacy, and emotional intelligence, alongside a heartwarming, cautious optimism. As one patient experimenting with a chatbot noted: \textit{"I could talk to the robot longer than I could talk to a human... she didn’t tell me if I’ve repeated myself... she [the bot] didn’t think I was boring".} \cite{xygkou2024mindtalker}

\vspace{-0.3em}
\paragraph{Overview} 
Figure~\ref{fig:dist_task_venue} shows the contributions of different scientific communities to the discussed task families, including literature reviews and dataset introductions. While all communities engage in dementia detection and dataset publication, caregiver support remains unexplored in NLP and Speech research. Additionally, we found no literature review similar to ours in NLP venues, highlighting a significant gap. The next section explores these and other gaps, highlighting opportunities for deeper, more rigorous research, as well as ethical dilemmas and innovative directions.


\section{Gaps and Opportunities}  
\label{sec:future}

\subsection{Data Variety}
\label{subsec:future_data}

\paragraph{Classic}
We begin our discussion of gaps by addressing a core aspect of NLP research: the data. As noted in Section \ref{sec:task_families}, the most popular datasets in the field are the Pitt corpus and its derivatives (ADReSS and ADReSSo), along with the CCC dataset (see additional details in Table~\ref{tab:datasets_summary}). These datasets are foundational and have significantly shaped the current state of NLP for dementia. However, like all data sources, they have limitations, including size constraints and demographic biases. For instance, the Pitt corpus contains more women than men, as well as an imbalanced ratio of 35\% healthy, 48\% dementia, and 17\% likely dementia participants. The ADReSS dataset was created as a balanced subset of Pitt, offering a smaller but evenly split sample by age, gender, and diagnosis \cite{vsevvcik2022systematic}. Additionally, these datasets capture only the language of individuals with the \textit{access, ability, and willingness to participate in clinical trials}. As a result, communities with less access to such trials, limited education, or intellectual and physical disabilities - factors known to impact cognitive assessments - are not represented \cite{bruhn2018assessment}. 

60\% of dementia patients live in low- and middle-income countries,\footnote{https://www.alzint.org/about/dementia-facts-figures/} making it crucial to gather data multilingual data from such communities. Some studies try to bridge this gap using translation, transfer learning, or cross-lingual statistical methods for resource-limited languages \citep{drame2012towards, fraser2019multilingual, lindsay2021language, guo2020text, perezautomatic, perez2022alzheimer, kabir2023early, meng2023integrated, melistascross}. One such example is \citet{nowenstein2024speech}, who applied translation for dementia detection in Icelandic. They caution that despite their promise, such methods need extensive validation due to varying grammatical complexities and dementia-related nuances across languages.

Another concern with these classic datasets is their timeliness. For instance, the Pitt Corpus, published in 1994, had a median participant age of 67. There is a significant difference between 67-year-olds now versus 30 years ago, including changes in technological proficiency and language use. This raises questions about whether models trained on these datasets are effective for detection in today’s elderly population, and whether these datasets may be reproduced with updated protocols.

Additionally, these classic datasets were manually transcribed. As such, they contain clean, standardized transcriptions with expert annotations of linguistic nuances such as disfluencies, pauses, and correction. However, manual transcription and annotation are neither scalable nor cost-effective. In real-world applications, texts will therefore likely be transcribed using automatic speech recognition (ASR), which produces noisier outputs and often fails to capture critical linguistic markers of cognitive decline \citep{al2018role, balagopalan2019impact, gomez2023alzheimer, heitz2024influence}. As a result, models trained on these clean, expert-transcribed datasets may struggle to generalize to real-world settings.

A final limitation of these classic datasets stems from the dementia stakeholder they represent- namely, the \textit{patient}. Other humans involved, such as caregivers, family members, and even the individuals conducting the cognitive assessments \cite{tahami2022stakeholder}, are not represented. Given the data-driven nature of NLP, this may naturally cause practitioners to focus on patients and inadvertently constrain the scope of research on all other dementia stakeholders.

\begin{table*}[ht!]
\centering
\begingroup
\normalsize
\renewcommand{\arraystretch}{1.5} 
\setlength{\tabcolsep}{12pt}      
\resizebox{1\linewidth}{!}{
\begin{tabular}{|l|l|c|l|l|c|l|}
\hline
\rowcolor[HTML]{FFFFC7} 
\cellcolor[HTML]{FFFFFF}                                        & \textbf{Name} & \textbf{Year} & \textbf{Type}                                                                  & \textbf{Population}    & \textbf{Modalities}                                                                                                   & \textbf{Size}        \\ \hline
\cellcolor[HTML]{DAFDFF}                                        & Pitt          & 1994          & Cognitive Assessments                                                          & Patients               & Transcribed speech                                                                                                    & 196 Dem vs. 98 Ctrl  \\ \cline{2-7} 
\cellcolor[HTML]{DAFDFF}                                        & ADReSS        & 2020          & Cognitive Assessments                                                          & Patients               & Transcribed speech                                                                                                    & 78 Dem vs. 78 Ctrl   \\ \cline{2-7} 
\cellcolor[HTML]{DAFDFF}                                        & ADReSSo       & 2021          & Cognitive Assessments                                                          & Patients               & Transcribed speech                                                                                                    & 121 Dem vs. 116 Ctrl \\ \cline{2-7} 
\multirow{-4}{*}{\cellcolor[HTML]{DAFDFF}\textbf{Classic}}      & CCC           & 2011          & \begin{tabular}[c]{@{}l@{}}Health and Wellbeing \\ Conversations\end{tabular}  & Patients               & Transcribed speech                                                                                                    & 125 Dem vs.125 Ctrl  \\ \Xhline{4\arrayrulewidth}
\cellcolor[HTML]{F9D9FE}                                        & CareD         & 2023          & Social media posts                                                             & Caregivers             & Caregivers written text                                                                                               & 1005 Posts           \\ \cline{2-7} 
\cellcolor[HTML]{F9D9FE}                                        & LoSST-AD      & 2024          & Public interviews                                                              & Patients (Celebrities) & Transcribed Interviews                                                                                                & 10 Dem vs. 10 Ctrl   \\ \cline{2-7} 
\multirow{-3}{*}{\cellcolor[HTML]{F9D9FE}\textbf{Contemporary}} & Li et-al      & 2023          & \begin{tabular}[c]{@{}l@{}}Clinical Notes +\\ Synthetic Sentences\end{tabular} & Clinicians / Patients  & \begin{tabular}[c]{@{}c@{}}(1) Human + Synthetic clinical notes \\ (2) Synthetically generated sentences\end{tabular} & 30k sentences        \\ \hline
\end{tabular}
}
\caption{\small Comparison between classic, widely-used datasets focused on transcribed interviews and cognitive assessments, and examples of contemporary datasets showcasing caregiver data, synthetic data, and more. For the full list of 17 datasets encountered throughout our review, see Appendix B.}
\vspace{-0.3em}
\label{tab:datasets_summary}
\endgroup
\end{table*}

\paragraph{Contemporary}

We now turn to data sources beyond clinical trials. \textit{Social media posts} have proven valuable for capturing the broad spectrum of communication across various dementia stakeholders (discussed in Section~\ref{sec:task_families}). Some studies creatively used \textit{non dementia-specific datasets, like IMDB}, leveraging their size and multilinguality or translation or embedding extraction for dementia-related tasks \citep{mirheidari2018detecting, li2019detecting}.

An additional direction is freely available data of \textit{public figures diagnosed with dementia}. \citet{petti2023generalizability, petti2023much} and \citet{asllani2023using} examine personal writings and spontaneous speech from well-known individuals with dementia, while \citet{petti2024losst} curated a corpus of a similar nature (mentioned in Table~\ref{sec:full_dataset_table}). This data source is unique in spanning celebrities of different cultures and origins, various types of data sources (interviews, written texts, etc.), and multiple modalities (video, audio, etc.). Additionally, it allows for tracking these public figures throughout their disease journey, exploring their individual progression. \citet{berisha2015tracking} and \citet{wang2020personalized} emphasize this in their detailed case studies on the linguistic decline of President Ronald Reagan throughout his political career.

LLMs offer another promising approach to data collection, annotation, and augmentation, with proven success in enhancing clinical notes \citep{liu2023leveraging, koga2024evaluating, latif2024evaluation}. In a unique study, \citet{li2023two} showed that LLMs can label clinical records, generate clinical data, and spontaneous speech, \textit{improving detection performance beyond that of purely human-created datasets}. The datasets they present are, to the best of our knowledge, the only dementia-related synthetic datasets currently available.


\vspace{-0.4em}
\paragraph{Future}

Historically, `NLP for dementia' was synonymous with dementia detection, and the classic datasets containing annotated clinical records have proven well-suited for detection classifiers. However, to advance dementia research in all tasks and frontiers, NLP requires data that is abundant, diverse, and reflective of the disease’s complexity. Table~\ref{tab:datasets_summary} compares the traditional clinical datasets with the more contemporary ones we discussed, representing a small part of the \textbf{17 datasets} we identified in our review (Table~\ref{tab:my-table} in Appendix~\ref{sec:full_dataset_table}). We hope these resources inspire researchers to pursue data from creative sources, covering diverse dementia stakeholders, multiple languages, genders, cultures, and educational backgrounds, designing NLP solutions that are accessible to all.

\vspace{-0.3em}
\subsection{Personalized Approaches}
\label{sub:personalized}

Every dementia patient experiences the disease progression uniquely, shaped by their individual characteristics and `distinct personal language model'. As each generation produces more accessible data, a 60-year-old today has likely generated significantly more text than a 60-year-old two decades ago. Given this wealth of textual data sourced from older adults, NLP may facilitate personalized approaches for managing, tracking, or even diagnosing dementia by building individualized language models that monitor cognition on a personal level.

NLP-based digital twins for dementia patients represent a promising frontier in healthcare. These virtual models replicate a patient’s cognitive and behavioral patterns through their language, enabling personalized care by monitoring cognitive decline and adjusting treatment plans based on simulated behaviors \cite{andargoli2024intelligent}. A notable example is \citet{hong2024conect}, who developed digital twins by fine-tuning an LLM on longitudinal data from the I-CONECT clinical trial,\footnote{https://www.i-conect.org/} using these twins to evaluate a patient support chatbot.

An intriguing, yet unexplored, research direction involves creating personalized LLM companions modeled after loved ones, as cognitive stimuli from familiar individuals can positively impact dementia patients \cite{woods2012remcare}. An LLM trained on family-curated data could engage patients with their personal history, family stories, and interests - such as favorite bands or recipes - potentially offering a non-clinical treatment. Given the vast amount of personal data generated today, this approach seems both feasible and effective. With these research directions, NLP has the potential to shape the future of personalized dementia treatments, both medical and alternative.

\vspace{-0.3em}
\subsection{Trust and Scientific Rigor}

A key contribution of our work is drawing comparisons between the medical and technological communities, particularly in scientific modeling. Medical research typically begins with a hypothesis, followed by study design and data collection for analysis. In contrast, data scientists often start with pre-collected data, retrospectively identifying patterns without necessarily defining a specific hypothesis. This conceptual gap may be critical in medical contexts. Is retrospective data relevant to future cases (e.g., will a model trained on the 1994 Pitt corpus capture the same signals in today’s dementia patients)? Do the results indicate correlation rather than causation? The medical community may be hesitant to adopt solutions they perceive as correlations derived from retrospectively collected, often small datasets \cite{el2023trustworthy}.

\begin{figure}[t]
    \centering
    \includegraphics[width=1\linewidth]{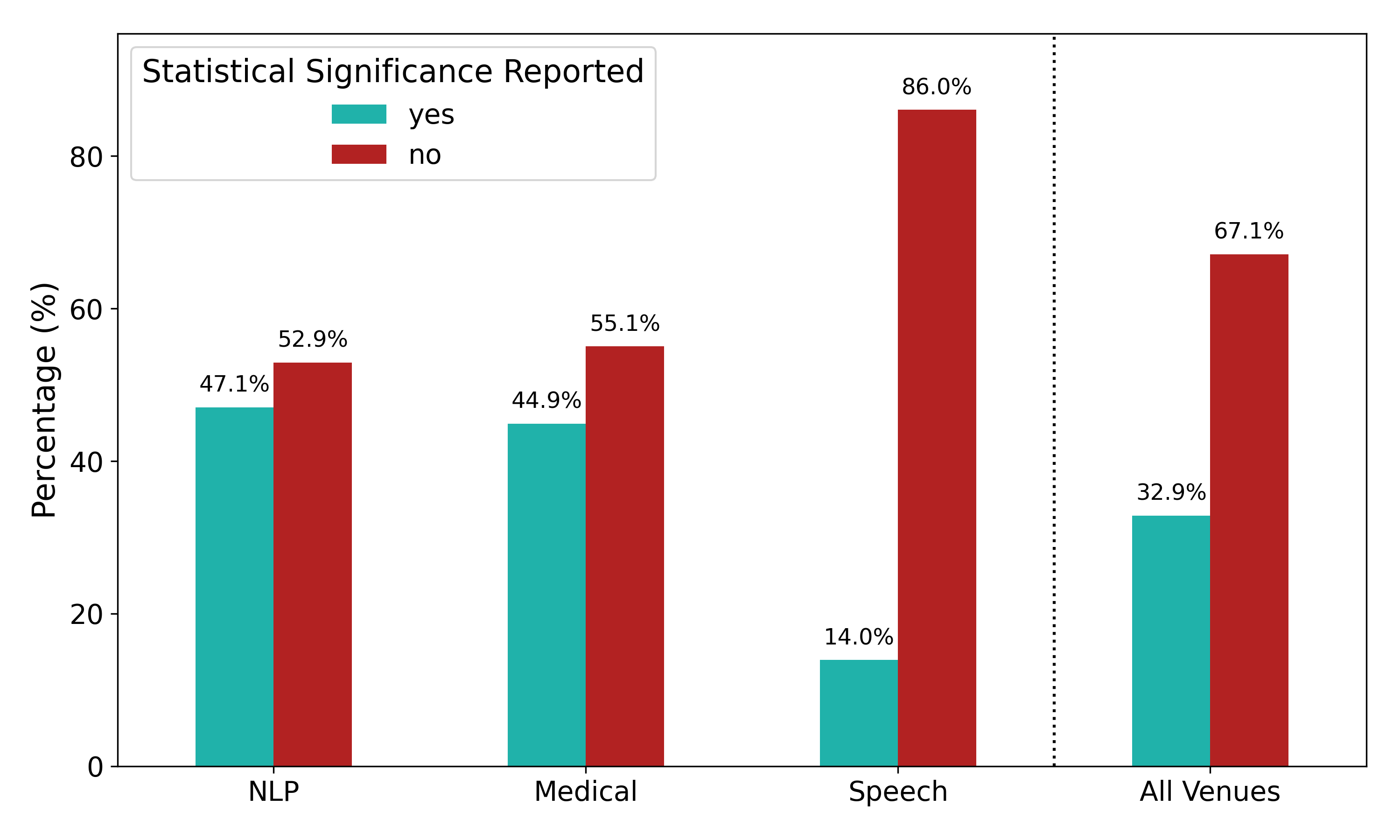}
    \captionsetup{skip=3pt}  
    \caption{Statistical significance per publication type.}
    \label{fig:stat_per_venue_type}
    \vspace{-0.8em}
\end{figure}

Hesitance to adopt NLP-based solutions may grow when studies do not report statistical significance or assess robustness. We manually annotated each paper in our cohort, excluding literature reviews, to indicate whether statistical tests or p-values were reported. Among all reviewed papers, \textbf{only 33\% report statistical significance}. While about half of NLP and medical papers include it, nearly 86\% of Speech publications - covering a significant portion of detection studies, particularly those using the popular ADReSS and ADReSSo datasets - do not (see Figure \ref{fig:stat_per_venue_type}).

Another crucial aspect of scientific modeling is robustness. We found few studies in our cohort which explicitly addressed this topic or evaluated the robustness of their models. Two notable examples are \citet{novikova2021robustness}, who examined BERT-based dementia detection under data perturbations, and \citet{favaro2024discovering}, who assessed the consistency of linguistic features across corpora.

Explainability, a key pillar of scientific rigor, remains rare in dementia detection, with few exceptions like \citet{ilias2022explainable}. A disconnect exists between the NLP community, which favors deep learning for higher accuracy, and the medical community, which prioritizes transparency and tends to distrust "black-box" models \citep{adadi2020explainable}. As a result, clinicians often prefer traditional, inherently interpretable models. Another gap exists as NLP practitioners’ favor feature-attribution methods, explaining prediction by assigning importance scores to individual tokens or features \citep{karlekar2018detecting, ilias2022explainable, vimbi2024interpreting}. While straightforward for researchers, these methods often produce overwhelmingly low-level explanations (e.g., assigning an importance score to every token in a lengthy transcript), potentially confusing both clinicians and patients, especially those with cognitive impairments. Addressing these challenges requires cross-disciplinary collaboration to develop interpretability methods that are both scientifically rigorous and clinically meaningful. We encourage researchers to consult surveys such as \citet{stiglic2020interpretability, calderon2024behalf, viswan2024explainable}, which outline NLP interpretability paradigms applicable across various algorithms.

Ensuring scientific rigor is just as crucial - and equally challenging - for tasks beyond dementia detection. Studies on caregiver support and patient assistance often lack a gold standard or clear objective. For example, evaluating LLMs that answer medical questions or simplify texts for dementia patients poses challenges in assessment and statistical rigor. Developing benchmarks to evaluate these tools is therefore a vital research direction.

\subsection{Application State-of-Mind}

To combat dementia in real-world settings, NLP practitioners must not only adopt rigorous methods and build trust- they should also consider the applicability of their work, and ask themselves - `how can my work be applied in practice?' Dementia detection exemplifies this gap- despite decades of promising results and hundreds of surveyed papers, our research found no evidence that these approaches have been integrated into real-world detection processes.\footnote{Based on publicly available data.} What hinders the clinical adoption of such promising NLP techniques? First, we believe that trust issues (as discussed earlier) are a major factor, as integrating these tools into medical workflows is too risky without rigorous testing across diverse datasets and real-world scenarios. However, they are not the only barrier.
 
One applicative challenge in dementia detection lies in the reliance on spoken responses, which must be transcribed in real time or post-recording. Clinicians often cannot record assessments due to privacy concerns and technological constraints, and even if recordings were conducted regularly, transcription would remain a bottleneck. Manual transcription is costly and time-consuming, making automatic speech recognition (ASR) the most practical alternative. Many studies have focused on improving ASR for dementia-related tasks - aligning transcriptions \citep{lehr2012fully, lehr2013discriminative}, tailoring ASR to specific dementia datasets \citep{abulimiti2020automatic}, and fine-tuning models like Whisper \citep{li2024whisper, radford2023robust}.Yet, ASR systems still introduce errors, miss key markers like disfluencies and repetition, and obscure the distinction between patient incoherence from transcription errors \cite{li2024useful}. NLP practitioners must consider the entire pipeline, not just their models. If solutions depend on error-prone ASR, they risk adding complexity rather than reducing burden.

Applicative thinking extends beyond dementia detection, relevant to all dementia-related tasks. \citet{topaz2020free}, \citet{zolnoori2023homeadscreen} and \citet{ryvicker2024using}, who tailor their NLP applications to the home healthcare setting, help professional caretakers in identifying, clustering, and analyzing dementia symptoms and behaviors (e.g., anxiety, delusions, hoarding). These systems were designed for real-world deployment, aiming to flag dementia-related signals in undiagnosed individuals and facilitate information transfer across healthcare settings, supporting both professional and personal caregivers. \citet{casu2024optimizing} present another form of applicative consideration, designing LLMs for dementia-related tasks and optimizing them for local fine-tuning to ensure data safety - an approach that might be more acceptable to the healthcare community compared to cloud-based training. 

Demand for dementia-related applications is evident from a business perspectives as well. The dementia treatment market is projected to surpass \$36 billion by 2030, with independent technological solutions rapidly emerging.\footnote{https://www.sphericalinsights.com/reports/dementia-drugs-market} For instance, the ChatGPT `app store' now features over 30 dementia-related chatbots for cognitive stimulation or virtual companionship.\footnote{Searched on https://chat.openai.com/, August 2024} People are taking initiative, leveraging LLMs to create these tools, and scientists can follow suit, using this framework to develop LLM-based playgrounds for large-scale applicative experiments. Such applications may also empower older adults to administer self-assesment and monitoring, as shown by \citet{skirrow2022validation}.

To summarize, the need for practical NLP solutions is undeniable, and we encourage the community to focus on solutions that are not only innovative but also truly applicable in real-life settings.

\begin{figure*}[h]
    \centering
    \includegraphics[width=1\linewidth]{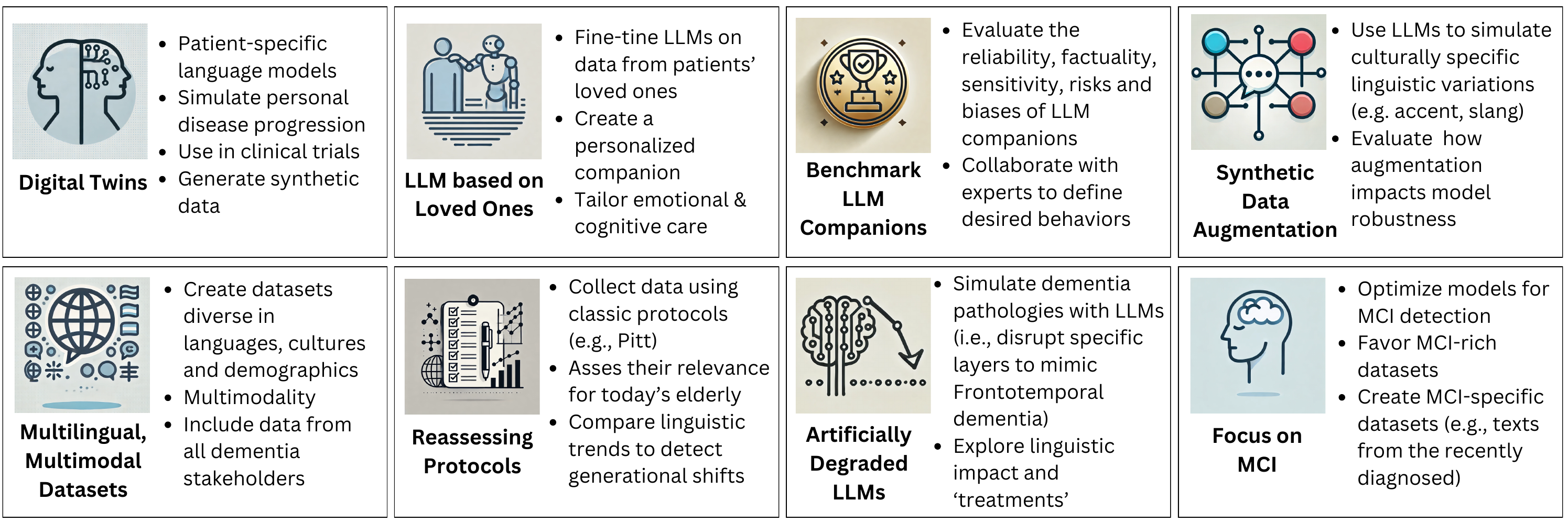}
    \caption{Key research directions we view as foundational, actionable, and yet-to-be fully explored.}
    \label{fig:key}
    \vspace{-1em}
\end{figure*}

\subsection{NLP Insights for Medicine}
\label{sub:med}

With many scientific communities advancing dementia research, cross-disciplinary collaboration is key. While NLP and other technological fields draw from medical insights and data, we argue that the reverse - integrating NLP insights into medical science and practice - holds immense potential.

One example is the concept of \textit{artificially degraded language models}, which explore `healthy' vs.\ `cognitively impaired' models replicating the linguistic behavior of dementia patients. \citet{cohen2020tale} achieved this by fine-tuning an LSTM on the DementiaBank cohort, while \citet{li2022gpt} degraded different layers of GPT-2. Both approaches yielded models that impressively demonstrated dementia-like linguistic patterns. The two studies utilized perplexity (a derivative of cross-entropy) as a measure of cognitive decline, a concept introduced in \citet{roark2011spoken}. The degraded models exhibit significantly lower perplexity when exposed to dementia-related linguistic anomalies compared to `healthy' models. We believe that this research area might shed light on the biological mechanisms of dementia and its pathologies- for example, degrading an LLM by mimicking protein deposits on its neurons might offer insights into the mechanisms of Alzheimer's.

\citet{li2024curious} use a degraded model to explore similarities between children's language and that of dementia patients. Their model shows lower perplexity than GPT-2 on picture descriptions by young children and dementia patients, and significantly higher perplexity on descriptions by children over 8, whose language aligns with that of with healthy adults. This work suggests NLP can aid in developing dementia biomarkers or cognitive assessments inspired by children's language.

NLP can also help craft novel cognitive assessments that examine more than the patient's performance of the task at hand. \citet{farzana2020modeling, nasreen2021rare, farzana2022interaction} and \citet{dawalatabad2022detecting} analyze interaction dynamics between patients and interviewers, demonstrating that rare dialogue acts, such as clarification exchanges and interviewer utterances, can signal cognitive variations. \citet{reeves2020narrative} propose an alternative assessment, focusing on narrative descriptions following video observations.

Another research direction can illuminate patients' lifestyle factors and experiences for the medical community. Studies such as \citet{yi2020natural, shen2022classifying, wu2023natural} and \citet{zhou2018comparison} focus on extracting patients' lifestyle factors from clinical records and social worker notes. They identify, for example, whether patients experience sleep deprivation, suffer from social isolation, or report substance abuse. Such NLP-based automated approaches can provide insights into a patient's unique experience of the disease and even allow for cognitive profiling \cite{upadhyay2022cognitive}, potentially allowing clinicians to tailor interventions that improve individualized patient outcomes.

Finally, NLP can prompt clinicians to reconsider existing hypotheses about dementia risk factors. \citet{zhou2019automatic, zhou2021identification} conducted extensive analyses of health records, uncovering contradictions with prevailing medical assumptions. While many medical studies emphasize cardiovascular risks - such as high-fat, high-calorie, and high-carbohydrate diets - these factors were rarely noted in dementia patient records. Instead, deficiencies in potassium, calcium, and other nutrients, as well as frequent mentions of dietary supplements (or their absence), appeared in over 25\% of dementia patients' records. Actual lab results for nutrient levels, however, were rarely documented, a result which challenges existing hypotheses, and exposes potential gaps in medical procedures. 


\section{Directions for Future Work}

Previous sections outlined the state of NLP for dementia and identified gaps and opportunities. Along the way, we highlighted open challenges that struck us as meaningful, practical, and yet to be fully explored (see Figure~\ref{fig:key}). We now distill these into concrete takeaways for researchers.

In Section~\ref{sub:personalized} we discussed personalized LLMs as a prominent direction. Namely, \textbf{digital twins} - personalized language models mimicking a patient's linguistic patterns - could help simulate disease progression, predict responses to clinical trials, or generate patient-specific data. Alternatively, \textbf{LLMs based on loved ones} - i.e., fine-tuned on texts from a patient’s family and friends - could enable truly personalized emotional and cognitive care. However, to ensure rigorous evaluation and adoption, \textbf{benchmarks for personalized LLMs} are a necessity. Researchers and domain experts must collaborate to create datasets, define ‘desired’ and ‘undesired’ behaviors, and establish standards for reliability, factuality, emotional capabilities, biases, risks, and ethical considerations (Section~\ref{sec:ethics}).

From a data perspective, we urge researchers to explore \textbf{synthetic data augmentation}- for example, making transcribed clinical assessments more representative of populations unable to attend (as discussed in Section~\ref{subsec:future_data}). Enhancing data with culturally diverse linguistic patterns, such as accents and regional slang, can help include these individuals without requiring direct trial participation.

Researchers should also prioritize \textbf{multilingual, multimodal datasets} to ensure dementia research worldwide reflects true linguistic, demographic, and cultural diversity. All stakeholders - patients, caregivers, clinicians, maybe even dementia-related content creators - should be represented. Another important direction is \textbf{reproducing classic protocols} like Pitt to interview older adults and collect data. This could evaluate their relevance, track generational language shifts, and test whether models trained on data from such protocols can generalize to today's population of older adults.

From a medical perspective, we believe that \textbf{artificially degraded language models} (discussed in Section~\ref{sub:med}) are the next frontier. Researchers should explore this by modeling pathologies with LLMs - disrupting specific model layers to mimic frontotemporal dementia or mimicking protein deposits on neurons to simulate Alzheimer’s. They can then analyze whether linguistic nuances emerge (e.g., does simulated frontotemporal dementia lead to more aggressive language?) and explore what might improve the LLM’s `condition.'

Finally, for NLP to truly support early detection, dementia must be diagnosed as early as possible. While it’s tempting to showcase high accuracy on Alzheimer's cases, this has limited clinical value since (a) clinicians can usually diagnose Alzheimer’s easily, and (b) few interventions exist at that stage. Researchers must therefore \textbf{focus on MCI}- develop algorithms attuned to the subtlest linguistic shifts and prioritize datasets with sufficient MCI cases - or even create MCI-specific datasets, such as texts from the recently diagnosed.

These research paths demand careful considerations, especially given the ethical complexities explored in the next section.

\section{Ethical Discussion}
\label{sec:ethics}

Ethical dilemmas regarding AI and healthcare have been widely examined \citep{rigby2019ethical, gerke2020ethical, keskinbora2019medical, harrer2023attention}. However, in this section, we highlight those specific to NLP and cognitive decline. One example is collecting data from the cognitively impaired- an ethically delicate task, as these individuals often cannot provide informed consent, raising questions about autonomy and data use. Fairness and diversity is another challenge, as NLP systems may misinterpret silent pauses, repetitions, or corrections, mistaking them for cognitive decline when they actually stem from non-native speech or even simply from stress.

The classic precision-recall trade-off also takes on new ethical weight in dementia research, particularly in detection. For treatable conditions like certain cancers, prioritizing recall to err on the side of caution is often justified. But for untreatable, progressive illnesses like dementia, false positives can be deeply harmful. A misdiagnosis, especially at a younger age, can lead to severe emotional distress, depression, and even suicidal thoughts \citep{lyketsos2003diagnosis, naismith2022suicidal}. Misdiagnosed individuals may even be prescribed newly emerging drugs and treatments, exposing them to serious side effects like brain edema and hemorrhages- despite not needing the medication. 

Another challenge may arise from using LLMs to generate patient speech. LLM hallucinations - non-factual outputs \cite{smith2023hallucination} - and confabulations - false memories without intent to deceive \cite{brown2017confabulation} - are problematic for many NLP tasks but are also key characteristics of dementia patients. How does one distinguish between LLM fabrications and authentic simulations of cognitive impairment? This raises both technical and ethical concerns: while some generated patterns (e.g., excessive empty speech or extreme hallucinations) may enhance accuracy, they can also introduce bias and reduce interpretability. Misleading data risks distorting our understanding of dementia, its cognition, and lived experience. Establishing benchmarks and ethical guidelines for synthetic data in this domain is essential.

Finally, researchers must ensure that models communicate predictions responsibly and safely. Dementia patients, often experiencing impaired judgment, may overshare sensitive information, act on unsafe advice, or follow suggestions with unintended consequences. For instance, if a chatbot recommends taking a walk to feel better, a patient might do so and become lost- an entirely plausible scenario. Addressing these ethical challenges demands cross-disciplinary collaboration, rigorous risk assessment, robust evaluation, and a strong sense of responsibility and accountability.

\section{Limitations}
Like all surveys, our work represents a snapshot in time. With NLP advancing extremely rapidly, new relevant studies are highly likely emerge after this review's publication. Additional limitations stem from deliberate choices to maintain a manageable scope and focused analysis. For instance, we relied on relatively broad dementia-related terms rather than expanding to all specific pathologies (e.g., Lewy Body, Vascular) or other potentially related medical conditions (e.g., Parkinson’s disease). Additionally, we focused on studies using data that is fully or partially in English, thus excluding research conducted solely in other languages. Finally, we confined our survey to peer-reviewed articles, acknowledging that in NLP, innovation often originates in open-source publications. Despite these limitations, we believe our review provides comprehensive coverage of the NLP for Dementia domain and provides a solid starting point for researchers looking to advance this field.

\section{Conclusions}

Through a cohort of 242 papers, we uncovered the depth and richness of NLP research for dementia. Multiple scientific communities are working to advance early detection, support patients and caregivers, and much more- all through language. Yet, many gaps and opportunities remain, and the data is out there. We urge researchers to collaborate, share knowledge through rigorous research, and uphold the highest ethical standards to make a true impact and help combat this disease.

\bibliography{NLP_for_Dementia}
\bibliographystyle{acl_natbib}


\appendix

\section{Cohort Construction}
\label{cohort}

\subsection{Search Methodology}
We searched in six data sources: ACL Anthology, PubMed, DBLP, IEEE Xplore, Springer, and Wiley. The database search was conducted between April and July 2024, with an additional search in January 2025 as part of our revised submission. Our research strategy focused on papers addressing dementia, Alzheimer's, and Mild Cognitive Impairment (MCI) using NLP, including but not limited to large language models (LLMs). 
We searched for papers including the keywords ("NLP" OR "Natural Language Processing" OR "LLM" OR "Language Model" OR "ChatGPT") AND ("Dementia" OR "Alzheimer" OR "MCI" OR "Cognitive Impairment") in their title, abstract, and/or keywords. While we do not include keywords for specific dementia pathologies (e.g., Lewy body), we believe that using the general term dementia in the title, abstract and/or keywords will effectively capture papers on various types of the condition.

\subsection{Paper inclusion and Exclusion Criteria}
\label{subsec:paper_inclusion_exclusion}

We screened the extracted articles against the following criteria: 
\begin{enumerate}[leftmargin=*,label=\alph*)]
    \itemsep=-3pt
    \item Only fully accessible, English-language papers were included.
    \item Only peer-reviewed papers were included.
    \item Only full academic papers (excluding posters and theses) were included.
    \item Studies focusing on text as the primary modality or employing multi-modal approaches that incorporate textual analysis were included. 
    \item Papers focusing solely on audio, visual, or other non-textual data were excluded.
    \item Papers that, despite mentioning our defined keywords, do not directly deal with NLP applications to Dementia, were excluded. 

\end{enumerate}

Papers were deemed irrelevant to our literature review if they contained dementia- and NLP-related terms in their title, abstract, or keywords but focused on entirely unrelated topics. For example, \citet{botros2020simple} mentions `dementia' in the abstract and `language model' in the keywords, but primarily focuses on sensor locations. Similarly, \citet{daudet2016portable} includes `Alzheimer’s' and `Language Model' as keywords but primarily discusses the design of an application for diagnosing brain injuries using youth speech data. 

As mentioned in Section~\ref{sec:methodology}, the first three steps of our screening process were automated, while the remaining steps - including gauging general relevance -  were conducted manually by two PhD students from our NLP lab, one of whom has a medical background in Alzheimer-related projects. We encountered discrepancies in fewer than 10\% of reviewed works, and resolved them through discussion until a final consensus was reached.

\begin{figure}[h]
    \centering
\includegraphics[width=1\linewidth]{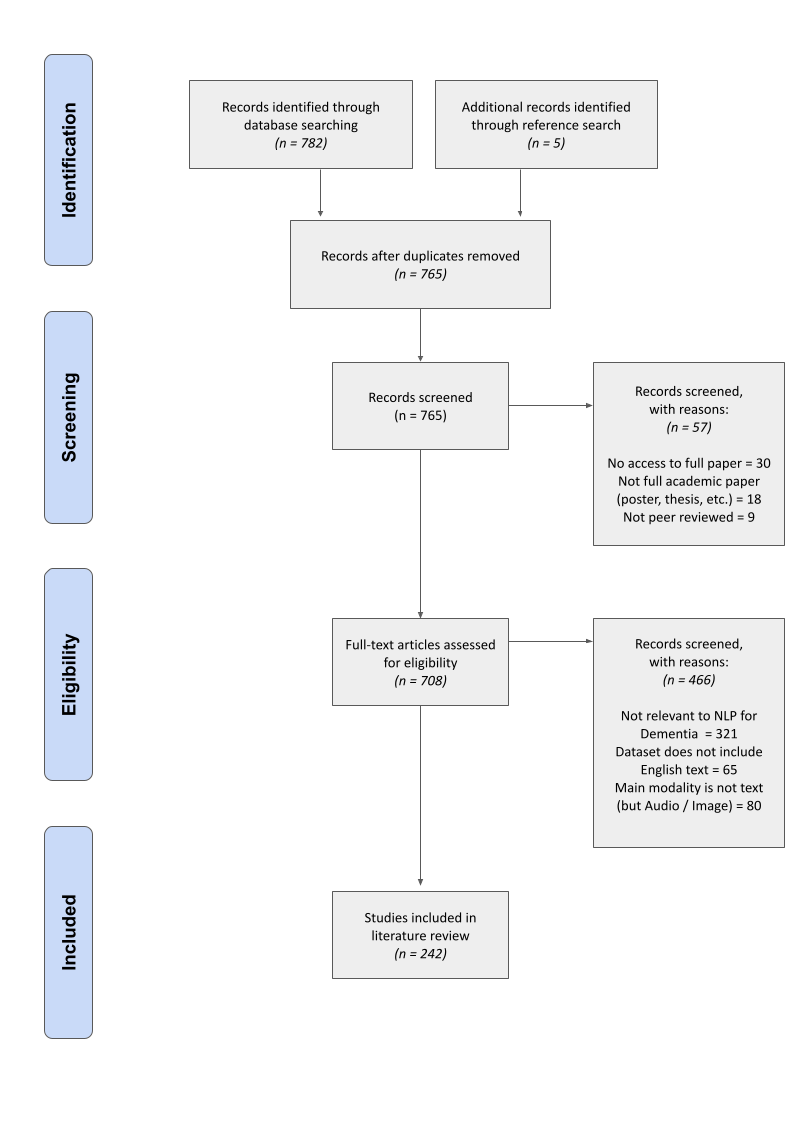}
    \caption{PRISMA flowchart displaying study screening and selection process.}
    \label{fig:prisma}
\end{figure}

A PRISMA flowchart describing our search process can be seen in  Figure~\ref{fig:prisma}.

\subsection{Dataset inclusion and Exclusion Criteria}
\label{subsec:dataset_inclusion_exclusion}

Throughout our review, we discuss classic and contemporary datasets, summarized in Table~\ref{sec:full_dataset_table}. The datasets mentioned in the paper and listed in this table were screened against the following criteria:

\begin{enumerate}[leftmargin=*,label=\alph*)]
    \itemsep=-3pt
    \item Only datasets referenced in the reviewed papers were included.
    \item The dataset must contain data that is fully or partially in English.
    \item Sufficient information about the dataset must be available online or obtainable through private requests to researchers or institutions.
    \item Datasets do not need to be publicly accessible to be included in our review.
    \item As these datasets are part of the peer-reviewed literature we selected, we assume they have also undergone peer review.
\end{enumerate}

\subsection{Paper Annotation}

For each paper, we extracted the title, authors, year of publication, and venue. We then manually annotated the following:
\begin{enumerate}[leftmargin=*,label=\alph*)]
    \itemsep=-3pt
    \item Task Family: Dementia detection, linguistic bio-marker extraction, caregiver support, patient assistance, literature review, or dataset introduction. While some papers may address multiple categories, we selected the most prominent motivation and novelty as described by the authors.
    \item Venue type: NLP (e.g., ACL), Medical (e.g., NIH), Speech (e.g.,Interspeech), or Other Technological (e.g., Frontiers in CS)
    \item Datasets used (if applicable).
    \item Technologies applied (if applicable), such as SVM or BERT.
    \item Statistical significance reported: yes or no, depending on whether any statistical tests were conducted on data or algorithmic results.
\end{enumerate}

\section{Extended Mention of Studies}
\label{extended_citations}

In Section~\ref{sec:task_families}, we cited a representative sample of dementia detection studies we reviewed. Below, we cite the remainder of the papers according to internal categorization used for our internal analysis. 

Detection papers using \textbf{classic ML classifiers} on a wide range of features, from straightforward N-grams and part-of-speech tagging to information content units extracted from picture descriptions and even dementia patients' dental records and motor signs: \citet{prud2011extraction, fraser2013using, jarrold2014aided, prud2015graph, fraser2016linguistic, yancheva2016vector, zhou2016speech, weissenbacher2016automatic, mirheidari2016diagnosing, santos2017enriching, wankerl2017n, klumpp2018ann, pou2018learning, eyre2020fantastic, chen2020topic, searle2020comparing, martinc2020tackling, hane2020predicting, al2021motor, nasreen2021alzheimer, clarke2021comparison, gonzalez2021automatic, soni2021using, penfold2022development, ablimit2022deep, dey2022integrated, amini2023automated, liu2023approach, taghibeyglou2023needs, wen2023revealing, patel2024utilizing, sharma2024detection}

Studies using classic neural models such as LSTMs and transformer-based models like BERT: \citet{karlekar2018detecting}, \citet{pompili2018topic}, \citet{pou2018learning}, 
\citet{hong2019novel}, 
\citet{di2019enriching}, \citet{chen2019attention}, 
\citet{fritsch2019automatic},
\citet{pan2019automatic},
\citet{edwards2020multiscale}, \citet{syed2020automated}, \citet{koo2020exploiting}, \citet{sarawgi2020multimodal}, \citet{pompili2020inesc}, \citet{cummins2020comparison}, \citet{balagopalan2020bert}, \citet{yuan2020disfluencies}, \citet{pappagari2020using}, \citet{haulcy2021classifying}, 
\citet{valsaraj2021alzheimer},
\citet{balagopalan2021comparing}, \citet{wang2021modular}, \citet{qiao2021alzheimer}, \citet{campbell2021alzheimer}, 
\cite{nambiar2022comparative}, 
\citet{khan2022stacked}, \citet{rohanian2021alzheimer}, \citet{syed2021tackling}, \citet{pappagari2021automatic}, \citet{perez2021influence}, \citet{mirheidari2021identifying}, \citet{liu2021automatic}, \citet{zhu2021exploring}, \citet{saltz2021dementia}, \citet{khan2022hsi}, \citet{deng2022alzheimer},
\citet{bouazizi2022dementia},
\citet{horigome2022identifying}, \citet{zheng2022evaluation}, \citet{liu2022improving}, \citet{liu2022transfer}, \citet{wang2022exploring}, \citet{matovsevic2022accurate}, \citet{amini2023automated}, 
\citet{rauniyar2023early}
\citet{abdelhalim2023training}, \citet{bouazizi2023dementia}, \citet{gkoumas2023reformulating}, \citet{cai2023multimodal}, \citet{wang2023exploiting}, \citet{shakeri2024uncovering}, \citet{dong2024hafformer}, \citet{kim2024machine}, \citet{laurentiev2024identifying}.

Studies classifying between more granular stages of the dementia (e.g.`MCI', `Alzheimer's'), assess disease progression or differentiate between dementia types (e.g., Alzheimer's versus Frontotemporal), through classification:
\citet{thomas2005automatic, orimaye2016deep, wang2020distinguishing, padhee2020predicting, merone2022multi, mao2023ad, liu2023harnessing, amini2024prediction, laurentiev2024identifying, panahi2024identifying}.
Studies focusing on text-based MMSE score regression for dementia detection: \citet{yancheva2015using}, \citet{pou2018learning}, \citet{farzana2020exploring}, \citet{rohanian2021multi}, \cite{stoppa2023graph},\citet{aryal2023ensembling}.

Studies presenting multi-modal approaches (with audio / video) as well as augmentation methods using synthetic, noisy, cross-domain, or semantically and structurally different datasets: \citet{che2017boosting, gligic2020named, gupta2020transfer, amin2020exploring, mahajan2021acoustic, guo2021crossing, zhu2021exploring, zhu2022domain, ablimit2022exploring, mirheidari2022automatic, hledikova2022data, ilias2022multimodal, el2022two, li2022alzheimer, maclagan2023can, chen2023assess, farzana2023towards, cui2023transferring, chenexploring, cai2023multimodal, lin2024multimodal}
\citet{fard2024linguistic}. 

Studies leveraging LLMs for detection or data augmentation: \citet{bullard2016towards}, \citet{liu2019roberta}, \citet{chintagunta2021medically}, \citet{liu2023leveraging}, \citet{cai2023multimodal}, \citet{duan2023cda}, \citet{koga2024evaluating}, \citet{latif2024evaluation}, \citet{casu2024optimizing}.

\section{Full Dataset Table}
\label{sec:full_dataset_table}

Table~\ref{tab:my-table} (presented in a separate page due to margin requirements of the TACL format) showcases the dementia-related datasets encountered throughout our review. Datasets such as the Reagan Library and IMDB, though creatively used in some papers, are excluded as they are not dementia-specific. Private health records or unreleased clinical notes are excluded as well.


\onecolumn
\begin{sidewaystable}[b!]
\centering
\scriptsize
\begin{tabular}{|l|lllll|llllll|ll|}
\hline
\multicolumn{1}{|c|}{\cellcolor[HTML]{DAFDFF}General}                                         & \multicolumn{5}{c|}{\cellcolor[HTML]{FFD07A}Data Source}                                                                                                                           & \multicolumn{6}{c|}{\cellcolor[HTML]{F9D9FE}Modalities}                                                                                                                                                              & \multicolumn{2}{c|}{\cellcolor[HTML]{DAE8FC}Participants}                                                                                                                                                                   \\ \hline
\textbf{Dataset}                                                                              & \multicolumn{1}{l|}{\textbf{Longit-}} & \multicolumn{1}{l|}{\textbf{Clinical}} & \multicolumn{1}{l|}{\textbf{Social}} & \multicolumn{1}{l|}{\textbf{Non-clinic.}} & \textbf{Synt.} & \multicolumn{1}{l|}{\textbf{Trans.}} & \multicolumn{1}{l|}{\textbf{Written}} & \multicolumn{1}{l|}{\textbf{Audio}} & \multicolumn{1}{l|}{\textbf{Video}} & \multicolumn{1}{l|}{\textbf{Medical}} & \textbf{Physical} & \multicolumn{1}{l|}{\textbf{Patients}}                                                                                & \textbf{Other}                                                                                      \\                                                                 & \multicolumn{1}{l|}{\textbf{udinal}}  & \multicolumn{1}{l|}{\textbf{setting}}  & \multicolumn{1}{l|}{\textbf{media}}  & \multicolumn{1}{l|}{\textbf{interviews}}  & \textbf{data}  & \multicolumn{1}{l|}{\textbf{speech}} & \multicolumn{1}{l|}{\textbf{text}}    & \multicolumn{1}{l|}{}               & \multicolumn{1}{l|}{}               & \multicolumn{1}{l|}{\textbf{imaging}} & \textbf{markers}  & \multicolumn{1}{l|}{}                                                                                                 &                                                                                                     \\ \hline
\begin{tabular}[c]{@{}l@{}}DementiaBank \\ \cite{becker1994natural}\end{tabular} & \multicolumn{1}{l|}{x}                & \multicolumn{1}{l|}{x}                 & \multicolumn{1}{l|}{-}               & \multicolumn{1}{l|}{-}                    & -              & \multicolumn{1}{l|}{x}               & \multicolumn{1}{l|}{-}                & \multicolumn{1}{l|}{x}              & \multicolumn{1}{l|}{-}              & \multicolumn{1}{l|}{-}                & -                 & \multicolumn{1}{l|}{\begin{tabular}[c]{@{}l@{}}196 Dem \\ 98 Ctrl\end{tabular}}                                       & -                                                                                                   \\ \hline
\begin{tabular}[c]{@{}l@{}}ADReSS \\ \cite{luz2021alzheimer}\end{tabular}        & \multicolumn{1}{l|}{-}                & \multicolumn{1}{l|}{x}                 & \multicolumn{1}{l|}{-}               & \multicolumn{1}{l|}{-}                    & -              & \multicolumn{1}{l|}{x}               & \multicolumn{1}{l|}{-}                & \multicolumn{1}{l|}{x}              & \multicolumn{1}{l|}{-}              & \multicolumn{1}{l|}{-}                & -                 & \multicolumn{1}{l|}{\begin{tabular}[c]{@{}l@{}}78 Dem \\ 78 Ctrl\end{tabular}}                                        & -                                                                                                   \\ \hline
\begin{tabular}[c]{@{}l@{}}ADReSSo \\ \cite{luz2021detecting}\end{tabular}       & \multicolumn{1}{l|}{-}                & \multicolumn{1}{l|}{x}                 & \multicolumn{1}{l|}{-}               & \multicolumn{1}{l|}{-}                    & -              & \multicolumn{1}{l|}{-}               & \multicolumn{1}{l|}{-}                & \multicolumn{1}{l|}{x}              & \multicolumn{1}{l|}{-}              & \multicolumn{1}{l|}{-}                & -                 & \multicolumn{1}{l|}{\begin{tabular}[c]{@{}l@{}}121 Dem \\ 116 Ctrl\end{tabular}}                                      & -                                                                                                   \\ \hline
\begin{tabular}[c]{@{}l@{}}CCC \\ \cite{pope2011finding}\end{tabular}            & \multicolumn{1}{l|}{x}                & \multicolumn{1}{l|}{-}                 & \multicolumn{1}{l|}{-}               & \multicolumn{1}{l|}{x}                    & -              & \multicolumn{1}{l|}{x}               & \multicolumn{1}{l|}{-}                & \multicolumn{1}{l|}{x}              & \multicolumn{1}{l|}{x}              & \multicolumn{1}{l|}{-}                & -                 & \multicolumn{1}{l|}{\begin{tabular}[c]{@{}l@{}}124 Dem \\ 125 Ctrl\end{tabular}}                                      & -                                                                                                   \\ \hline
\begin{tabular}[c]{@{}l@{}}B-SHARP \\ \cite{li2020analysis}\end{tabular}         & \multicolumn{1}{l|}{-}                & \multicolumn{1}{l|}{x}                 & \multicolumn{1}{l|}{-}               & \multicolumn{1}{l|}{-}                    & -              & \multicolumn{1}{l|}{x}               & \multicolumn{1}{l|}{-}                & \multicolumn{1}{l|}{x}              & \multicolumn{1}{l|}{-}              & \multicolumn{1}{l|}{-}                & -                 & \multicolumn{1}{l|}{\begin{tabular}[c]{@{}l@{}}144 Dem \\ 185 Ctrl\end{tabular}}                                           & -                                                                                                   \\ \hline
\begin{tabular}[c]{@{}l@{}}IVA \\ \cite{pan2020acoustic}\end{tabular}            & \multicolumn{1}{l|}{-}                 & \multicolumn{1}{l|}{x}               & \multicolumn{1}{l|}{-}               & \multicolumn{1}{l|}{-}                  & -              & \multicolumn{1}{l|}{-}             & \multicolumn{1}{l|}{-}              & \multicolumn{1}{l|}{x}            & \multicolumn{1}{l|}{-}              & \multicolumn{1}{l|}{-}              & -               & \multicolumn{1}{l|}{\begin{tabular}[c]{@{}l@{}}45 Dem \\ 25 Ctrl\end{tabular}}                                        & -                                                                                                   \\ \hline
Farmington Heart Study                                                                        & \multicolumn{1}{l|}{x}                & \multicolumn{1}{l|}{x}                 & \multicolumn{1}{l|}{-}               & \multicolumn{1}{l|}{-}                    & -              & \multicolumn{1}{l|}{x}               & \multicolumn{1}{l|}{x}                & \multicolumn{1}{l|}{x}              & \multicolumn{1}{l|}{-}              & \multicolumn{1}{l|}{x}                & x                 & \multicolumn{1}{l|}{\begin{tabular}[c]{@{}l@{}}\textgreater 15,000 over\\ decades\end{tabular}}            & -                                                                                                   \\ \hline
\begin{tabular}[c]{@{}l@{}}CareD \\ \cite{garg2023cared}\end{tabular}            & \multicolumn{1}{l|}{-}                & \multicolumn{1}{l|}{-}                 & \multicolumn{1}{l|}{x}               & \multicolumn{1}{l|}{-}                    & -              & \multicolumn{1}{l|}{-}               & \multicolumn{1}{l|}{x}                & \multicolumn{1}{l|}{-}              & \multicolumn{1}{l|}{-}              & \multicolumn{1}{l|}{-}                & -                 & \multicolumn{1}{l|}{-}                                                                                                & \begin{tabular}[c]{@{}l@{}}1005 caregiver\\ posts\end{tabular}                                               \\ \hline
\begin{tabular}[c]{@{}l@{}}ADNI \\ \cite{petersen2010alzheimer}\end{tabular}     & \multicolumn{1}{l|}{x}                & \multicolumn{1}{l|}{x}                 & \multicolumn{1}{l|}{-}               & \multicolumn{1}{l|}{-}                    & -              & \multicolumn{1}{l|}{-}               & \multicolumn{1}{l|}{x}                & \multicolumn{1}{l|}{-}              & \multicolumn{1}{l|}{-}              & \multicolumn{1}{l|}{x}                & x                 & \multicolumn{1}{l|}{\begin{tabular}[c]{@{}l@{}}\textgreater 2500 over \\ several cohorts\end{tabular}} & -                                                                                                   \\ \hline
\begin{tabular}[c]{@{}l@{}}Wisconsin WLS \\ \cite{herd2014cohort}\end{tabular}   & \multicolumn{1}{l|}{x}                & \multicolumn{1}{l|}{x}                 & \multicolumn{1}{l|}{-}               & \multicolumn{1}{l|}{x}                    & -              & \multicolumn{1}{l|}{x}               & \multicolumn{1}{l|}{x}                & \multicolumn{1}{l|}{-}              & \multicolumn{1}{l|}{-}              & \multicolumn{1}{l|}{-}                & x                 & \multicolumn{1}{l|}{\begin{tabular}[c]{@{}l@{}}\textgreater 10,000 over \\ 60 years\end{tabular}}      & -                                                                                                   \\ \hline
\begin{tabular}[c]{@{}l@{}}MCSA\\ \cite{roberts2008mayo}\end{tabular}            & \multicolumn{1}{l|}{x}                & \multicolumn{1}{l|}{x}                 & \multicolumn{1}{l|}{-}               & \multicolumn{1}{l|}{-}                    & -              & \multicolumn{1}{l|}{-}               & \multicolumn{1}{l|}{x}                & \multicolumn{1}{l|}{-}              & \multicolumn{1}{l|}{-}              & \multicolumn{1}{l|}{x}                & x                 & \multicolumn{1}{l|}{\textgreater  3000}                                                                 & -                                                                                                   \\ \hline
\begin{tabular}[c]{@{}l@{}}Layton Aging and Alz. \\ Disease Research Center\end{tabular}      & \multicolumn{1}{l|}{-}                & \multicolumn{1}{l|}{x}                 & \multicolumn{1}{l|}{-}               & \multicolumn{1}{l|}{x}                    & -              & \multicolumn{1}{l|}{x}               & \multicolumn{1}{l|}{-}                & \multicolumn{1}{l|}{x}              & \multicolumn{1}{l|}{x}              & \multicolumn{1}{l|}{x}                & x                 & \multicolumn{1}{l|}{\begin{tabular}[c]{@{}l@{}}Depending \\ Sub-cohort\end{tabular}}                                  &                                                                                                     \\ \hline
I-CONECT                                                                                      & \multicolumn{1}{l|}{x}                & \multicolumn{1}{l|}{-}                 & \multicolumn{1}{l|}{-}               & \multicolumn{1}{l|}{x}                    & -              & \multicolumn{1}{l|}{x}               & \multicolumn{1}{l|}{-}                & \multicolumn{1}{l|}{x}              & \multicolumn{1}{l|}{x}              & \multicolumn{1}{l|}{-}                & -                 & \multicolumn{1}{l|}{320*}                                                                                             & -                                                                                                   \\ \hline
\begin{tabular}[c]{@{}l@{}}LoSST-AD \\ \cite{petti2024losst}\end{tabular}        & \multicolumn{1}{l|}{x}                & \multicolumn{1}{l|}{-}                 & \multicolumn{1}{l|}{-}               & \multicolumn{1}{l|}{x}                    & -              & \multicolumn{1}{l|}{x}               & \multicolumn{1}{l|}{-}                & \multicolumn{1}{l|}{-}              & \multicolumn{1}{l|}{-}              & \multicolumn{1}{l|}{-}                & -                 & \multicolumn{1}{l|}{\begin{tabular}[c]{@{}l@{}}10 Dem \\ 10 Ctrl\end{tabular}}                                        & -                                                                                                   \\ \hline
\begin{tabular}[c]{@{}l@{}}SLaCAD \\ \cite{farzana2024slacad}\end{tabular}       & \multicolumn{1}{l|}{x}                & \multicolumn{1}{l|}{x}                 & \multicolumn{1}{l|}{-}               & \multicolumn{1}{l|}{-}                    & -              & \multicolumn{1}{l|}{x}               & \multicolumn{1}{l|}{-}                & \multicolumn{1}{l|}{x}              & \multicolumn{1}{l|}{-}              & \multicolumn{1}{l|}{-}                & x                 & \multicolumn{1}{l|}{\begin{tabular}[c]{@{}l@{}}9 Dem \\ 82 Ctrl\end{tabular}}                                         & -                                                                                                   \\ \hline
\citet{li2023two}                                                                & \multicolumn{1}{l|}{-}                & \multicolumn{1}{l|}{x}                 & \multicolumn{1}{l|}{-}               & \multicolumn{1}{l|}{-}                    & x              & \multicolumn{1}{l|}{-}               & \multicolumn{1}{l|}{x}                & \multicolumn{1}{l|}{-}              & \multicolumn{1}{l|}{-}              & \multicolumn{1}{l|}{-}                & -                 & \multicolumn{1}{l|}{-}                                                                                                & \begin{tabular}[c]{@{}l@{}}(1) Clinicians \\ (2) Synt. Annotated \\ (3) Synt. Generated \\ 16,000 sentences each.\end{tabular} \\ \hline
\citet{gkoumas2024longitudinal}                                                  & \multicolumn{1}{l|}{x}                & \multicolumn{1}{l|}{-}                 & \multicolumn{1}{l|}{-}               & \multicolumn{1}{l|}{x}                    & -              & \multicolumn{1}{l|}{x}               & \multicolumn{1}{l|}{x}                & \multicolumn{1}{l|}{x}              & \multicolumn{1}{l|}{-}              & \multicolumn{1}{l|}{-}                & -                 & \multicolumn{1}{l|}{\begin{tabular}[c]{@{}l@{}}12 Dem \\ 10 Ctrl\end{tabular}}                                        & -                                                                                                   \\ \hline
\end{tabular}
\captionsetup{width=0.8\textwidth}
\caption{Overview of datasets reviewed. Most datasets originate from transcribed or written speech in clinical settings, with quite a few offering longitudinal data. Two under-represented categories here are synthetic datasets and social media-based datasets, with only one of each category. For additional sources of social media data we refer the readers to \citet{sunmoo2023comparing} and \citet{masrani2017detecting}-while they do not publish full datasets, they provide code to replicate their scraping process. For non-English datasets, see \citet{yang2022deep}. \\  ** Estimated number of participants in the I-CONECT clinical trial are based on publicly available data.}
\label{tab:my-table}
\end{sidewaystable}


\end{document}